\documentclass[letterpaper]{article} 
\usepackage{aaai2026}  
\usepackage{times}  
\usepackage{helvet}  
\usepackage{courier}  
\usepackage[hyphens]{url}  
\usepackage{graphicx} 
\usepackage{natbib}  
\usepackage{caption} 
\usepackage{multirow} 
\usepackage{pifont} 
\usepackage{amsmath}
\usepackage{bm}
\usepackage{amsfonts}
\usepackage{placeins}
\usepackage{siunitx}
\newcommand{\figref}[1]{\textbf{Figure~\ref{#1}}}
\newcommand{\tabref}[1]{\textbf{Table~\ref{#1}}}
\newcommand{\secref}[1]{\textbf{Section~\ref{#1}}}
\newcommand{\eqnref}[1]{\textbf{Equation~\ref{#1}}}

\usepackage{algorithm}
\usepackage{algorithmic}

\usepackage{newfloat}
\usepackage{listings}
\DeclareCaptionStyle{ruled}{labelfont=normalfont,labelsep=colon,strut=off} 
\floatstyle{ruled}
\newfloat{listing}{tb}{lst}{}
\floatname{listing}{Listing}
\title{Activation-Wise Propagation: A One-Timestep Strategy \\for Spiking Neural Networks}
\author{
    Jian Song\textsuperscript{\rm 1},
    Xiangfei Yang\textsuperscript{\rm 2}\footnotemark[1],
    Shangke Lyu\textsuperscript{\rm 3},
    Donglin Wang\textsuperscript{\rm 1}\thanks{Corresponding author.}
}

\affiliations{
    \textsuperscript{\rm 1}Westlake University\\
    \textsuperscript{\rm 2}Hangzhou City University\\
    \textsuperscript{\rm 3}Nanjing University\\
    songjian76@westlake.edu.cn, yxf9011@163.com, shangke\_lyu@nju.edu.cn, wangdonglin@westlake.edu.cn
}

\begin{document}

\maketitle

\begin{abstract}
Spiking neural networks (SNNs) have demonstrated significant potential in real-time multi-sensor perception tasks due to their event-driven and parameter-efficient characteristics. A key challenge is the timestep-wise iterative update of neuronal hidden states (membrane potentials), which complicates the trade-off between accuracy and latency. SNNs tend to achieve better performance with longer timesteps, inevitably resulting in higher computational overhead and latency compared to artificial neural networks (ANNs). Moreover, many recent advances in SNNs rely on architecture-specific optimizations, which, while effective with fewer timesteps, often limit generalizability and scalability across modalities and models. To address these limitations, we propose Activation-wise Membrane Potential Propagation (AMP2), a unified hidden state update mechanism for SNNs. Inspired by the spatial propagation of membrane potentials in biological neurons, AMP2 enables dynamic transmission of membrane potentials among spatially adjacent neurons, facilitating spatiotemporal integration and cooperative dynamics of hidden states, thereby improving efficiency and accuracy while reducing reliance on extended temporal updates. This simple yet effective strategy significantly enhances SNN performance across various architectures, including MLPs and CNNs for point cloud and event-based data. Furthermore, ablation studies integrating AMP2 into Transformer-based SNNs for classification tasks demonstrate its potential as a general-purpose and efficient solution for spiking neural networks.
\end{abstract}



\section{Introduction}

Due to the low power consumption and the emergence of neuromorphic sensors, Spiking Neural Networks (SNNs) have gained attention to address perception tasks. Unlike artificial nonlinear activation functions such as ReLU or sigmoid, SNNs encode representations using discrete spikes, mimicking the behavior of biological neurons. Taking the Leaky Integrate-and-Fire (LIF) neuron as an example, the spiking process consists of three stages (\eqnref{lif}--\eqnref{lif3}): accumulation , emission, and reset. A hidden state known as membrane potential (MP) is used to regulate both historical and current information by decaying past MP while integrating incoming inputs. 
\begin{align}
    U_{i}(t) &= H_{i}(t-1) + X_{i}(t) \label{lif}\\
    S_{i}(t) &= Hea.(U_{i}(t)-V_{th}) \label{lif2}\\
    H_{i}(t) &= \beta U_{i}(t)(1-S_{i}(t)) + V_{th}S_{i}(t)\label{lif3}
\end{align}
Here, $i$ and $t$ denote the layer and timestep, $U_{i}(t)$ and $H_{i}(t)$ represent the MP of neurons before and after activation at timestep $t$. $V_{th}$ represents the threshold for spike generation. $Hea.(\cdot)$ denotes the Heaviside step function defined as $Hea.(x) = 1$ if $x \geq 0$, and $Hea.(x) = 0$ otherwise. The leaky mechanism depends on the decay factor $\beta $, which attenuates the influence of historical information.

\begin{figure}[!hbt]
\centering
\includegraphics[width=\columnwidth]{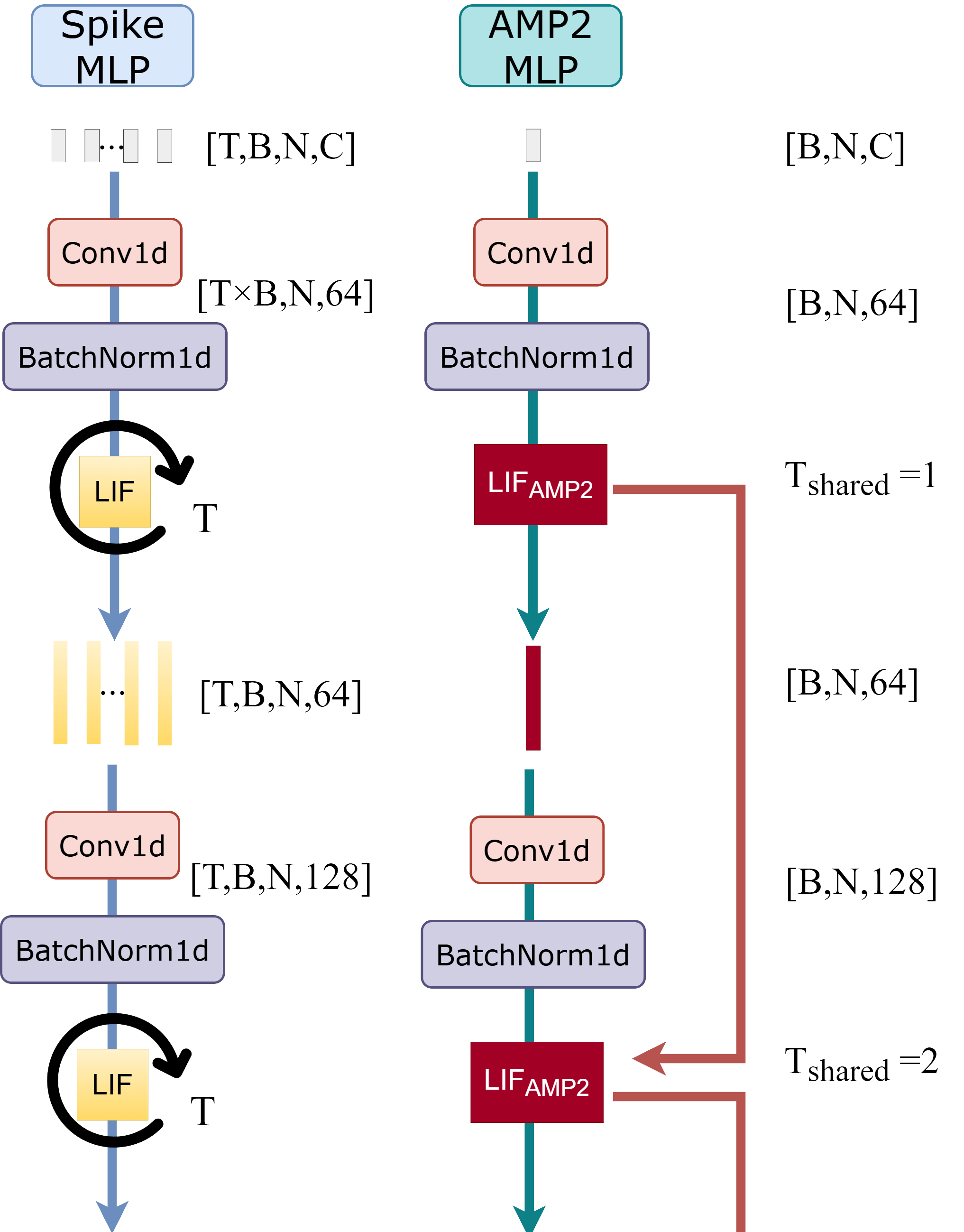}
\caption{Forward propagation of Spiking PointNet. Left: \textbf{T}imestep-\textbf{W}ise \textbf{U}pdate (TWU). Right: \textbf{A}ctivation-wise \textbf{M}embrane \textbf{P}otential \textbf{P}ropagation (AMP2). AMP2 replaces timestep-wise iterative MP accumulation with neuron-wise accumulation, enabling an SNN that operates effectively at a single timestep.}
\label{fig:teaser}
\end{figure}

SNNs have increasingly matched artificial neural networks (ANNs) in performance and efficiency, particularly narrowing the gap in training efficiency for static images. The Spike-Driven Transformer \cite{yao2023spikedriven, yao2024spikedriven, yao2024scaling} continually optimizes the spiking mechanism within the transformer architecture, utilizing only 4 timesteps for static images and 16 timesteps for neuromorphic data captured by dynamic vision sensors (DVS). Spikformer \cite{zhou2023spikformer, zhou2024spikformer} introduces spiking self-attention and a convolution-attention hybrid mechanism, outperforming other ResNet-based SNNs on both static and DVS images. Notably, QKFormer \cite{zhou2024qkformer} trained with 4 timesteps, surpasses several ANNs on ImageNet-1k for the first time. 

For neuromorphic data, a common pre-processing involves integrating an event episode into a number of frames, which are required to match the timesteps of SNN. To preserve fine-grained information, researchers typically integrate a larger number of frames (often 10 or 16). This requirement leads to longer timesteps for SNN when processing DVS data compared to static images, further reducing the training efficiency of SNNs. Alternatively, some studies have proposed treating events as 3D event clouds \cite{wang2019space} to circumvent the integration step. Although SpikePoint \cite{ren2023spikepoint} performs well on multiple DVS benchmarks, it still relies on the 16-timestep training. Therefore, reducing the number of timesteps remains a key challenge for SNNs.

Such performance has not yet been achieved in the 3D point cloud domain. \citet{wu2025spiking} proposed Spiking Point Transformer (SPT), which exploited spiking self-attention for Point Transformer \cite{zhao2021point}. With increased model size, SPT readily outperforms the Spiking PointNet \cite{ren2023spiking}. Similarly, \citet{wu2024point} scaled up convolutional SNN by stacking 13 spiking residual blocks in P2SResLNet, drawing upon KPConv \cite{thomas2019kpconv}. Despite these advances in scaling, there remains a lack of \textbf{lightweight yet effective} methods for SNN on point clouds, whereas some techniques have demonstrated effectiveness for SNNs on 2D static images (discussed in \secref{sec:shrinking}).

\begin{table}[!bt]
    \centering
    \begin{tabular}{clll}
    \hline
        Timestep & Memory per GPU & GPU Hours & Acc.(\%)\\ 
        \hline
        1 & 30.15GB & 42.96 &78.0\\ 
        4 & 74.87GB & 98.64 &80.36\\ 
    \hline
    \end{tabular}
    \caption{Training efficiency of Spikformer(8-512) on ImageNet100 with a batch size of 120 under different timesteps.}
    \label{tab:spikformer}
\end{table}

The iterative \textbf{T}imestep-\textbf{W}ise  \textbf{U}pdate (TWU, shown on the left of \figref{fig:teaser}) of neuronal MP remains the predominant approach in current SNN models. For static RGB images, these methods typically replicate the input along an additional temporal dimension to enable TWU. Specifically, an image input with shape $[B, C, H, W]$ is augmented with a temporal dimension $T$, resulting in input of shape $[T, B, C, H, W]$, where $B$, $C$, $H$, and $W$ denote batch size, channels, height, and width, respectively. $T$ determines the number of intermediate spike emissions, i.e., the number of iterative calculations for MP.  A large $T$ ensures stable spiking representations. However, this pre-processing introduces data redundancy, leading to excessive memory consumption and prolonged spiking iterations. \tabref{tab:spikformer} shows that increasing timesteps from \textbf{1 to 4} results in a $\boldsymbol{2.5\times}$ increase in \textbf{memory} consumption and a $\boldsymbol{2.3\times}$ increase in training \textbf{time}.

When $T$ is small, most spiking neurons may  not reach the firing threshold, resulting in a narrow and underdeveloped MP distribution. Consequently, the generated spikes may inadequately represent the underlying information. Furthermore, TWU inevitably amplifies the variance of MP distribution during temporal iterations, regardless of the value of $T$, due to the variance accumulation of hidden states. To address the instability of SNN outputs, \citet{ding2025rethinking} proposed MP smoothing and guidance, aiming to reduce inter-timestep variability in MP distribution. While this approach demonstrates effectiveness under moderate timesteps, it does not fundamentally resolve the issues of computational burden and latency caused by temporal iterations. This naturally raises the question of \textbf{whether it is possible to achieve high performance while eliminating timestep-wise iterative updates}, that is, an effective single-timestep MP update. Information accumulation through neighboring neurons in the human brain may be a solution. 

Unlike conventional SNNs, where iterative MP updates and forward propagation alternate, we propose a parallel propagation scheme for both MP and spikes, eliminating the need for timestep-wise MP updates. This method, termed \textbf{A}ctivation-wise \textbf{M}embrane \textbf{P}otential \textbf{P}ropagation (AMP2) is illustrated in the right of \figref{fig:teaser}. The main contributions of this paper are summarized as follows:

\begin{itemize}
    \item AMP2 consistently improves SNN performance without iterative updates ($T=1$) across multiple architectures, including Spiking PointNet, Spiking ResNet, and Spiking Transformers.
    \item AMP2-based PointNet series outperform low-timestep baselines with more parameters on ModelNet40 \cite{wu20153d} and ScanObjectNN\cite{uy2019revisiting}.
    \item We conduct extensive experiments on multiple datasets, covering classification and semantic segmentation of point clouds, frame-based and point-based recognition on neuromorphic data, as well as static image classification on ImageNet100 \cite{imagenet100_kaggle}. 
\end{itemize}

\section{Related Works on Reducing Timesteps}

\label{sec:shrinking}
Timestep reduction is an emerging area. \citet{ding2024shrinking} proposed Shrinking SNN (SSNN), which divides SNN training into multiple stages with progressively decreasing timesteps. SSNN employs larger timesteps in the early stages and gradually reduces them in later stages. Each stage involves a complete classifier training, ultimately achieving strong performance with average timesteps of 5 for DVS frames. \citet{ding2025rethinking} identified excessive differences in MP distributions across timesteps as a significant factor disrupting SNN training. They demonstrate that smoothing and guidance on temporally adjacent MP effectively alleviate such differences at 5 timesteps. \citet{zuo2024temporal} introduced temporal reversal augmentation (TRR) to regularize SNN training. By reversing the temporal order of inputs or features, TRR enhances SNN robustness to varying timesteps. As a result, timesteps for neuromorphic and static datasets are further reduced to 5 and 2, respectively. Notably, TRR demonstrates compatibility with existing models including VGG-9, MS-ResNet34 \cite{MS_ResNet}, Spike-Driven Transformer \cite{yao2023spikedriven}, and PointNet series \cite{Qi_2017_CVPR,qi2017pointnet++}. Spiking Point Transformer and P2SResLNet compensate for reduced timesteps by employing larger model architectures; specific parameter counts are provided in \textbf{Appendix}.

\section{Method}

\subsection{Overall Framework}
\begin{figure*}[!tb]
    \centering
    \includegraphics[width=0.95\textwidth]{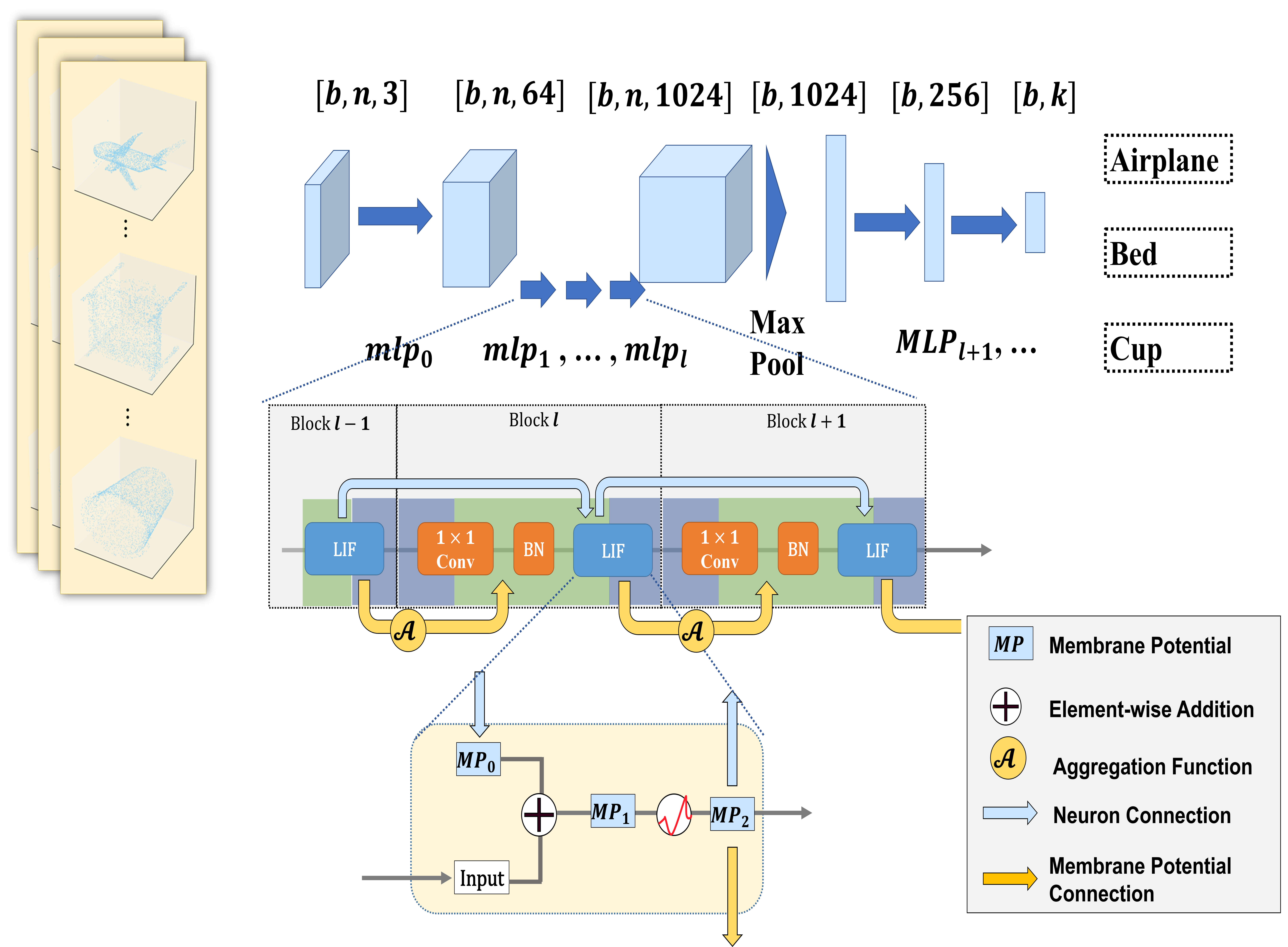}
    \caption{The workflow of Activation-wise Membrane Potential Propagation (AMP2) applied in Spiking PointNet.}
    \label{fig:overview}
\end{figure*}

\textbf{A}ctivation-wise \textbf{M}embrane \textbf{P}otential \textbf{P}ropagation (AMP2) is explained based on the LIF neuron described in \eqnref{lif}--\eqnref{lif3}. To provide an intuitive illustration of AMP2, we  use simplified notations $MP^0$, $MP^1$, and $MP^2$ to represent AMP2's MP at the initialization, accumulation, and reset stages, respectively, distinguishing them from $H(t)$ and $U(t)$, which denote MP in accumulation and reset stages in conventional TWU neurons. We introduce two key novelties to the TWU of the LIF neuron, shown in \figref{fig:overview} and pseudocode in \textbf{Appendix}.
\begin{itemize}
    \item Activation/Neuron-\textbf{W}ise \textbf{P}ropagation (AWP): We explicitly enable direct propagation of MP between neurons, in parallel with the forward propagation of features. The initial $MP^0$ is primarily determined by the residual membrane potential $MP^1$ carried over from the previous spiking layer, rather than being initialized to a fixed value or random perturbation \cite{ren2023spiking}.
    \item Residual Learning of MP (RMP): Residual learning helps mitigate spike degradation, especially in deep SNNs. Inspired by existing residual connection in SNNs, we introduce a novel residual pathway between MP and features. 
\end{itemize}

\subsection{Activation/Neuron-wise Propagation (AWP)}
For a fair comparison, we adopt a surrogate gradient function for LIF neuron as used in Spiking PointNet \cite{ren2023spiking}. \\
\textbf{Forward Propagation:}
\begin{align}
    MP_{random} &\sim \mathcal{U}(0, c), \quad \text{where } c \in (0, 1]
\\
    MP^0_i &=\alpha MP_{i-1}^{1} + (1-\alpha)MP_{random}\\
    MP^1_i &= \beta MP^0_i + X_i  \label{eqn:beta} \\
    S_i &= Hea.(\frac{MP^1_i}{V_{th}} - c) \\
    MP^2_i &=MP_i^1(1-S_i)
\end{align}
\textbf{Surrogate Gradient Function:}
\begin{align}
    x_i &= \frac{MP^1_i}{V_{th}} \label{eqn:x}\\
    g(x_i) &= \frac{tanh(k(x_i-c))+tanh(kc)}{2tanh(kc)}\\
    \frac{\partial g(x_i)}{\partial x_i} &= \frac{k \cdot \mathrm{sech}^2(k(x_i - c))}{2\tanh(kc)}
\end{align}
$k$ represents the scaling factor that controls the $\tanh$ function; larger values of $k$ make the surrogate gradient approach a hard boundary. $c$ denotes the implicit threshold in the surrogate gradient function. The gradient of the loss with respect to the weights, $\frac{\partial \mathcal{L}_{CE}}{\partial W_i}$, is directly proportional to $\frac{\partial \mathcal{L}_{CE}}{\partial MP^0_i}$ \cite{ren2023spiking, ding2025rethinking}, which is given by
\begin{align}
     \frac{\partial \mathcal{L}_{CE}}{\partial MP^0_i} &= \frac{\partial L}{\partial S_i} \cdot \frac{\partial S_i}{\partial MP^1_i} \cdot \frac{\partial MP^1_i}{\partial MP^0_i} \label{eqn:loss}\\ 
      \frac{\partial S_i}{\partial MP^1_i} &\approx \frac{\partial g(x_i)}{\partial x_i} \cdot \frac{1}{V_{th}} \\
      \frac{\partial \mathcal{L}_{CE}}{\partial MP^0_i}&= \frac{\partial \mathcal{L}_{CE}}{\partial S_i} \cdot \frac{k \cdot \mathrm{sech}^2(k(x_i - c))}{2\tanh(kc)} \cdot \frac{\beta}{V_{th}} \label{eqn:back}
\end{align}
According to \eqnref{eqn:back},we can find that $k$, $c$, $\beta$, $V_{th}$ are constants. The partial derivative $\frac{\partial \mathcal{L}_{CE}}{\partial S_i}$ determines the contribution of spike $S_i$ to the overall loss gradient. 

if $x_i$ deviates significantly from $c$, $\mathrm{sech}^2(\cdot)$ would approach zero, causing the overall gradient to vanish during backpropagation. Conversely, if $x_i$ is extremely close to $c$ and $kc$ is extremely small, the gradient may become excessively large:
\begin{align*}
    tanh(kc) &\approx kc \; \text{if}\;kc \ll 1\;, and\; x \rightarrow c\\
     \implies g\prime(x_i) & \approx \frac{k\cdot \mathrm{sech}^2(0)}{2kc} = \frac{1}{2c}
\end{align*}
Gradients may explode as $c \rightarrow 0$. Therefore, it is important to keep $c$ and $k$ within appropriate ranges.

We demonstrate the effectiveness of AMP2 in the context of MP initialization. Since $x_i$ depends on $MP_i^1$ (\eqnref{eqn:x}), we compare AMP2 with two other prevailing strategies for $MP^0_i$:

(1) Zero initialization (Pure Input: $MP_i^0=0$)\\
This commonly used approach completely ignores residual MP between spatially adjacent
spiking neurons. Compensation is typically achieved through timestep-wise accumulation. Since $X_i$ is produced by a normalization layer, its expectation tends to remain constant and nonadjustable.
\begin{align*}
    MP^1_i &= X_i\\ 
    \mathbb{E}[x_i] &= \mathbb{E}[\frac{\beta MP^0_i+X_i}{V_{th}}]=\frac{\mathbb{E}[X_i]}{V_{th}} \approx 0 \\
    \mathrm{Var}(x_i) &= \frac{\mathrm{Var}(X_i)}{V_{th}^2} \to \frac{1}{V_{th}^2} 
\end{align*}

(2) Random initialization (Pure Perturbation: $\alpha =0$ )\\
\citet{ren2023spiking} proposed this method to enhance generalization by introducing perturbations to MP. While such perturbations do benefit Spiking PointNet, the method also relies on increasing timesteps to maintain this effect. The updating efficiency of all gradients is adjusted by $\beta$.
\begin{align*}
     MP^1_i &= \beta MP_{random}^i + X_i\\ 
    \mathbb{E}[x_i] &= \frac{\beta \mathbb{E}[MP_{random}] + \mathbb{E}[X_i]}{V_{th}} \approx \frac{\beta c}{2V_{th}}\\
    \mathrm{Var}(x_i) &= \frac{1}{V_{th}^2} \left[ \beta^2 \mathrm{Var}(MP_{random}) + \mathrm{Var}(X_i) \right] \\
    &\to \frac{1}{V_{th}^2} \left(  \frac{\beta^2c^2}{12} + 1 \right)
\end{align*}

(3) AMP2 (Memory-Perturbation Fusion)\\
From a neuroscience perspective, spatially proximate neurons can influence each other's states through molecular diffusion, even without direct synaptic connections. This phenomenon can be observed in AMP2 that $MP^1_i$ would integrate input $X_{i-k}$ from adjacent neurons. Rather than random policy aiming for uniform gradient updates across depth, AMP2 encourages depth-related gradient updates. Furthermore, although we replace timestep-wise accumulation with spatially adjacent accumulation to save computational resources, AMP2 remains fundamentally compatible with both approaches. This can be interpreted as a global neuron leveraging an ensemble effect. 
\begin{align*}
    & MP^1_i = \sum_{k=0}^{i-1} (\beta\alpha )^k \left[ \beta (1-\alpha )MP_{random}^{i-k} + X_{i-k} \right] + (\beta\alpha)^i MP_0^1\\ 
    &\mathbb{E}[MP^1_i] \approx \beta (1-\alpha)\frac{c}{2} \cdot \frac{1-(\beta\alpha)^{i}}{1-\beta\alpha}\\
    & \mathbb{E}[x_i] = \frac{\mathbb{E}[MP^1_i]}{V_{th}} = \frac{\beta(1-\alpha)\frac{c}{2}}{V_{th}} \cdot \frac{1-(\beta\alpha)^{i}}{1-\beta\alpha}\\
    &(\beta \alpha)^i \to 0\ \text{when $\boldsymbol i$ increases}\\
     &\lim_{i \to \infty} \mathbb{E}[x_i] = \lim_{i \to \infty} \frac{\mathbb{E}[MP^1_i]}{V_{th}} =  \frac{\beta c(1-\alpha)}{2V_{th}(1-\beta\alpha)}\\
\end{align*}
\subsection{Residual MP Connection (RMP)}

\begin{table}[!tp]
    \centering
    \begin{tabular}{cll}
        \hline
        & Type & Targets\\
        \hline
          A & Vanilla  & $f(Spike, Features)$\\
          B & Spike-Element-Wise  & $f(Spike, Spike)$\\
          C & Membrane  & $f(Features, Features)$\\
          D & AMP2  & $f(MP, Features)$\\ 
        \hline
    \end{tabular}
    \caption{Definition of three existing residual connections (A-C) in SNNs and AMP2 (D).}
    \label{tab:residual}
\end{table}
\begin{figure}[!bt]
\centering
\includegraphics[width=0.9\columnwidth]{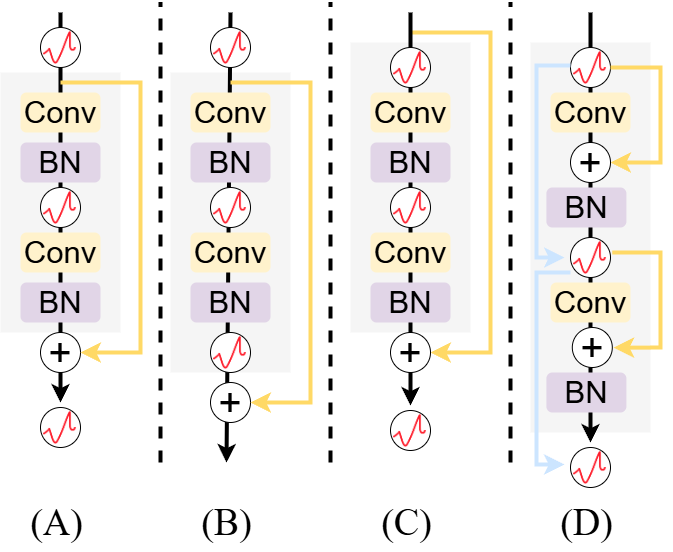}
\caption{Illustration of existing shortcuts (A-C) in SNNs and AMP2 (D). Yellow lines indicate residual connection paths, while blue line denotes activation-wise propagation.}
\label{fig:residual}
\end{figure}
We compare three existing residual shortcuts (A, B, C) with the one used in AMP2 (D) in \tabref{tab:residual} and \figref{fig:residual}. \citet{zheng2021going} introduced a vanilla shortcut between spikes and normalized features, mimicking the original residual connection. While this preserves the spike-driven nature, it lacks the identity mapping function. \citet{fang2021deep} presented an alternative path for pure spikes; this SEW approach transforms spike-driven characteristics into integer-driven ones. Subsequently, \citet{MS_ResNet} exploited the Membrane Shortcut to address previous limitations. Although this shortcut is claimed to connect MP, it actually integrates normalized features that have not been accumulated as MP. In contrast, we propose a novel approach that establishes a genuine connection between membrane potentials.
\subsection{Complexity and Energy Efficiency}
\begin{figure}[!bt]
    \centering
    \includegraphics[width=0.9\columnwidth]{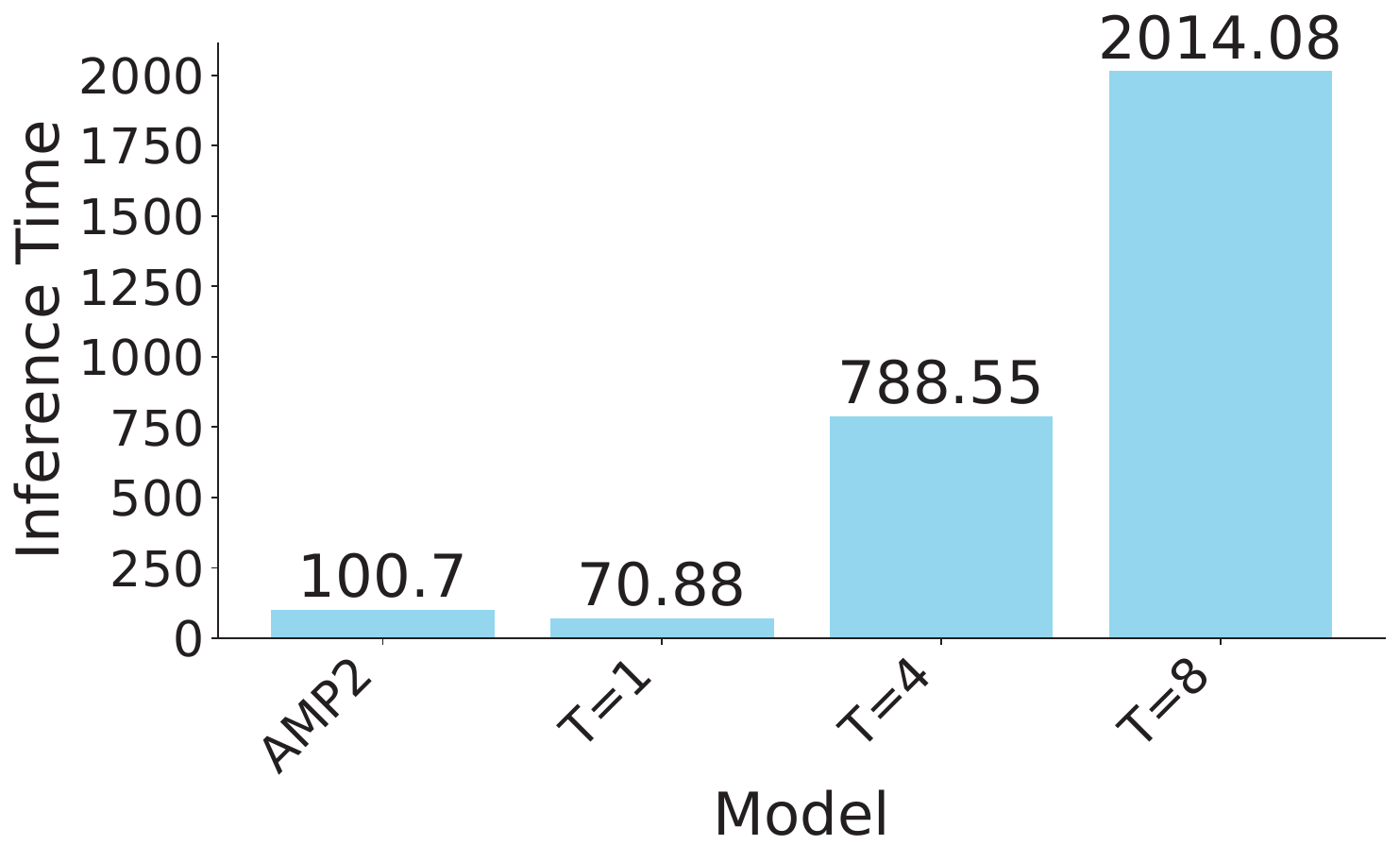}
    \caption{Impact of timesteps ($1,4,8$) on the inference time (seconds) of Spiking PointNet for a single batch, evaluated on the neuromorphic hardware Lynxi HP201.}
    \label{fig:inference}
\end{figure}

\begin{table}[!tb]
  \centering
  \begin{tabular}{cccc}
    \hline
    Activation & AMP2&Space  &  Time \\
    \hline
    ReLU& \ding{55}& $O(BCN)$ & $O(BCN)$\\
    \multirow{2}{*}{LIF}& \ding{55}& $O(2TBCN)$  & $O(TBCN)$\\
     & \ding{51}&$O(2BCN)$ &$O(BCN)$\\
  \hline
\end{tabular}
  \caption{Complexity comparison of ReLU, timestep-based LIF and activation-based LIF}
  \label{tab:complex}
\end{table}
Energy consumption in SNNs is primarily dominated by accumulate operations (AC) and multiply-accumulate operations (MAC). According to measurements by \citet{horowitz20141}, a single 32-bit floating-point MAC consumes \SI{4.6}{\pico\joule} ($E_{MAC}$), while an AC operation consumes only \SI{0.9}{\pico\joule} ($E_{AC}$). Based on \tabref{tab:complex}, the energy consumption of standard LIF processing can be estimated by \eqnref{eqn:std_lif}, whereas the consumption of AMP2 depends on the specific element-wise functions $f(\cdot)$. The power costs of $ADD$ and $AND$ are estimated by \eqnref{eqn:add} and \eqnref{eqn:and}, respectively. It is evident that AMP2 reduces the computational overhead compared to the commonly used 4 or 8 timesteps. A more intuitive comparison on inference latency can be seen in \figref{fig:inference}.

\begin{align}
    E_{LIF} & = T B  C  N \cdot E_{MAC} \label{eqn:std_lif}\\
    E_{AMP2(+)} &= B  C  N \cdot (E_{AC} + 2\cdot E_{MAC}) \label{eqn:add}\\
    E_{AMP2(\times)} &= 3  B C N \cdot E_{MAC} \label{eqn:and}
\end{align}

\section{Experiment}
\subsection{3D Point Cloud Recognition}
\begin{table*}[!htbp]
\centering
    \begin{tabular}{lllllllllll}
        \hline
        \multirow{2}{*}{Arch.} & \multirow{2}{*}{Source} &  \multirow{2}{*}{Method} & \multirow{2}{*}{Type} & \multirow{2}{*}{T($\downarrow$)} & \multicolumn{2}{c}{ModelNet40($\uparrow$)} & \multicolumn{2}{c}{ScanObjectNN($\uparrow$)} &  \\ 
         &  &     &  &  & OA & mACC & OA& mACC\\ 
         \hline
        \multirow{6}{*}{PointNet} & \citet{Qi_2017_CVPR}   & PointNet$^\dagger$ & ANN & 1 & 90.43 & 86.66 & 68.40 & 62.38 \\ 
         & \multirow{3}{*}{\citet{ren2023spiking}}   & \multirow{3}{*}{Spiking PointNet$^\dagger$} & SNN & 1 & 87.36 & 82.58 & 24.84 & 26.51 \\ 
         &  &     & SNN & 4 & 88.41 & 83.43 & 64.04$^\ddagger$ & 60.14$^\ddagger$ \\ 
         &  &    & SNN & 8 & 88.50 & 84.73 & - & - \\ 
         & \citet{lan2023efficient} & ANN2SNN & SNN & 16 & 88.17 & - & \textbf{66.56} & - \\ 
         & \citet{zuo2024temporal}& TRR & SNN & 2 & 88.84&-&-&-\\
         & - &   AMP2(Ours) & SNN & 1 & \textbf{89.74} & \textbf{85.60} & 65.48 & \textbf{60.29} \\ 
         \hline
        \multirow{7}{*}{PointNet++(SSG)} & \citet{qi2017pointnet++}   & PointNet++$^\dagger$ & ANN & 1 & 92.33  & 90.07 & 84.46 & 82.62 \\ 
         & \multirow{3}{*}{\citet{ren2023spiking}}  &   \multirow{3}{*}{Spiking PointNet++$^\dagger$} & SNN & 1 & 87.36 & 82.58 & 48.01 & 43.77 \\ 
         &  &    & SNN & 4 & 91.22 & 87.43 & - & - \\ 
         &  &    & SNN & 8 & 91.61 & 88.59 & - & - \\ 
         & \citet{lan2023efficient}  & ANN2SNN & SNN & 16 & 89.45 & - & 69.22 & - \\ 
         & \citet{ding2025rethinking}  & MP Smooth+Guidance & SNN & 2 & 91.13 & - & - & - \\ 
         & \multirow{2}{*}{\citet{zuo2024temporal}}& TRR & SNN & 1 & 89.65& - &- &-\\
         & & TRR & SNN & 2 &90.57&-&-&-\\
         & - &  AMP2(Ours) & SNN & 1 & \textbf{91.23} & \textbf{88.99} & 72.57 & 67.78 \\ 
         PointNet++(MSG) & - & AMP2(Ours) & SNN & 1 & \textbf{91.63} & \textbf{89.16}& \textbf{78.81} & 75.71\\
         \hline
         Efficient 3D SNN & \citet{qiu2024efficient} & E3DSNN & SNN & 1& 91.50 & -& - & -\\
         \hline
        \multirow{2}{*}{KPConv} & \multirow{2}{*}{\citet{wu2024point}} & P2SResLNet-B & SNN & 1 & 90.60 & 89.20 & 74.46$^\ddagger$ & 72.58$^\ddagger$\\
         & & KPConv-SNN & SNN & 40 & 70.5 & 67.6 & 43.90 & 38.70 \\ 
        
         \hline
        \multirow{3}{*}{Point Transformer} & \multirow{3}{*}{\citet{wu2025spiking}}& \multirow{3}{*}{SPT(Q-SDE768)} & SNN & 1 & 90.87 & - & 76.33 & - \\ 
        & & & SNN & 2 & 91.13 & 88.93 & 77.03 & - \\ 
         &   &  & SNN & 4 & 91.22 & 88.45 & 78.03 & \textbf{75.87} \\
        \hline  
    \end{tabular}
    \caption{Classification accuracy of existing SNNs at low timesteps on ModelNet40 and ScanObjectNN. The best results for each metric are highlighted in \textbf{bold}. $\dagger$ indicates self-reproduced results, while $\ddagger$ denotes results reproduced by \citet{wu2025spiking}. T indicates timesteps used in SNNs.}
    \label{tab:cls}
\end{table*}
From \tabref{tab:cls}, we can see that AMP2 enables both Spiking PointNet and Spiking PointNet++ to achieve superior performance at a single timestep on ModelNet40 and ScanObjectNN. Notably, AMP2 not only allows SNNs to outperform conventional counterparts training with 4 timesteps, but also surpasses other low-timestep approaches, such as TRR and MP Smooth+Guidance. Moreover, AMP2 facilitates lightweight SNNs in achieving marginal outperformance over significantly larger baseline models, including E3DSNN, P2SResLNet, and Spiking Point Transformer. AMP2-based Spiking PointNet++ with multi-scale grouping (MSG) demonstrates higher overall accuracy than single-scale (SSG), achieving 91.63\% on ModelNet40 and 78.81\% on ScanObjectNN, which represents the best results among point cloud SNNs to date.

\begin{table}[!tb]
    \centering
    \begin{tabular}{lllllll}
    \hline
        \multirow{2}{*}{Type} & \multirow{2}{*}{AMP2}& \multirow{2}{*}{T} & \multicolumn{2}{c}{SharpNetPart($\uparrow$)}&   \multicolumn{2}{c}{S3DIS($\uparrow$)}  \\ 
        &  & & Instance & Class & Point & Class \\ 
        \hline
        ANN& \ding{55}& 1 & 84.48 & 80.61 & 79.59 & 54.38 \\ 
        SNN &\ding{55}& 1 & 78.46 & 72.76 & 70.66 & 46.10 \\ 
         SNN &\ding{55}& 4 & 81.38 & 75.53 & 72.42 & \textbf{47.79} \\ 
        SNN &\ding{51}& 1 & \textbf{82.44} & \textbf{78.31} & \textbf{74.21} & 46.70 \\ 
        \hline
    \end{tabular}
    \caption{Effect of AMP2 on Spiking PointNet for semantic segmentation task of point cloud}
    \label{tab:seg}
\end{table}

We assessed the effectiveness of AMP2 in improving the 3D semantic segmentation performance of Spiking PointNet on two datasets: ShapeNetPart \cite{10.1145/2980179.2980238} and the Stanford 3D Indoor Scene \cite{armeni20163d} Dataset (S3DIS). As presented in \tabref{tab:seg}, AMP2 enables Spiking PointNet to achieve superior mean intersection over union (mIoU) for both instance and category segmentation compared to the 4-timestep model. Additionally, on the S3DIS dataset, AMP2 surpasses the 4-timestep SNN in point recognition accuracy.

\subsection{Neuromorphic Recognition}
From \tabref{tab:dvs128}, we evaluate AMP2 on neuromorphic data DVS128Gesture \cite{8100264} using both frame-based and point-based inputs. For frame input, AMP2 is combined with timestep-wise updating on the ResNet-tiny architecture (7B-Net) proposed by \citet{fang2021deep}. Consequently, AMP2 consistently provides stable improvements for three variant SNNs with 4 timesteps. 

\begin{table*}[htbp]
    \centering
    \begin{tabular}{lllllcll}
    \hline
        Input & Arch. & Method & Type & Source  & AMP2 & T($\downarrow$) & Acc. \\ \hline
        \multirow{14}{*}{Frame} & \multirow{4}{*}{VGG-9} & SSNN & \multirow{4}{*}{SNN} & \citet{ding2024shrinking}$^{AAAI}$& \multirow{4}{*}{\ding{55}} & \multirow{4}{*}{5} & 90.74 \\ 
         &  & TRR &  & \citet{zuo2024temporal}& &  & 91.67 \\ 
         &  & SLT &  & \citet{anumasa2024enhancing}$^{AAAI}$& &  & 89.35 \\ 
         &  & MP Smooth+Guidance &  & \citet{ding2025rethinking}$^{NeurIPS}$ & &  & 93.23 \\
         \cline{2-8}
        ~ & \multirow{6}{*}{ResNet-tiny} & \multirow{2}{*}{SEWResNet$^{\dagger}$} & \multirow{6}{*}{SNN} & \multirow{6}{*}{\citet{fang2021deep}$^{NeurIPS}$} &\ding{55} & \multirow{6}{*}{4} & 88.19 \\ 
         & ~ &  &  & & \ding{51} &  & \textbf{93.75(+5.56)} \\ 
         & ~ & \multirow{2}{*}{Spiking ResNet$^{\dagger}$} &  & & \ding{55} &  & 71.88 \\ 
         &  &  &  & &\ding{51} &  & \textbf{82.29(+10.41)} \\ 
         &  & \multirow{2}{*}{Spiking PlainNet$^{\dagger}$} & & & \ding{55} &  & 75.0 \\ 
         &  &  & & & \ding{51} &  & \textbf{78.13(+3.13)} \\ 
         \cline{2-8}
         & \multirow{4}{*}{Transformer} & SpikingResformer-Ti & \multirow{4}{*}{SNN} & \citet{shi2024spikingresformer}$^{CVPR}$ &\multirow{4}{*}{\ding{55}} & 5 & 90.63 \\ 
         & & Spikformer & & \citet{ding2024shrinking}$^{AAAI}$ & & 5 & 79.52\\
         &  & SDT-v1(2-256) & & \citet{ding2025rethinking}$^{NeurIPS}$& & 5 & 92.24 \\ 
         & & SDT-v1(2-256) & & \citet{yao2023spikedriven}$^{NeurIPS}$& & 16 & 99.30 \\
         \hline
        \multirow{11}{*}{Point} & \multirow{5}{*}{PointNet} & PointNet$^{\dagger}$ & ANN & \citet{Qi_2017_CVPR}$^{CVPR}$& \ding{55} & 1 & 89.57 \\ 
         &  & Space-time Event Cloud & ANN & \citet{wang2019space}$^{WACV}$ & \ding{55} & 1 & 88.77 \\ 
         &  & \multirow{2}{*}{Spiking PointNet$^{\dagger}$} & SNN & \multirow{2}{*}{\citet{ren2023spiking}$^{NeurIPS}$}&\ding{55} & 1 & 31.22 \\ 
         &  &  & SNN & & \ding{55} & 4 & 91.32 \\ 
         & ~ & AMP2(Ours) & SNN & - & \ding{51} & 1 & \textbf{92.82(+61.6)} \\ 
         \cline{2-8}
        ~ & \multirow{5}{*}{PointNet++} & PointNet++$^{\dagger}$ & ANN & \citet{qi2017pointnet++}$^{NeurIPS}$& \ding{55} & 1 & 96.74 \\ 
         &  & Space-time Event & ANN & \citet{wang2019space}$^{WACV}$ &\ding{55} & 5 & 95.70 \\ 
         &  & \multirow{2}{*}{Spiking PointNet++$^{\dagger}$} & SNN & \multirow{2}{*}{\citet{ren2023spiking}$^{NeurIPS}$}& \ding{55} & 1 & 85.0 \\ 
         &  &  & SNN & & \ding{55} & 4 & 94.68 \\ 
         &  & AMP2(Ours) & SNN & - & \ding{51} & 1 & \textbf{92.89(+7.89)} \\ 
         \cline{2-8}
         & SpikePoint & SpikePoint & SNN & \citet{ren2023spikepoint}$^{ICLR}$& \ding{55} & 16 & \textbf{98.74} \\ \hline
    \end{tabular}
    \caption{Comparative results (\%) of existing low-timesteps SNNs on DVS128Gesture. $\dagger$ represents self-reproduced experiments.}
    \label{tab:dvs128}
\end{table*}

\section{Ablation Study}

\subsection{Influence of Network Depth}

\begin{table}[!htb]
    \centering
    \begin{tabular}{llccc}
    \hline
        \multirow{2}{*}{SA} & \multirow{2}{*}{Param(M)} & \multicolumn{2}{c}{AMP2}& \multirow{2}{*}{Acc.(\%)} \\ 
         &  & AWP & RMP &  \\ 
        \hline
        \multirow{4}{*}{3} & \multirow{4}{*}{1.4757} & \ding{55} & \ding{55} & 87.35 \\ 
         &  & \ding{55} & \ding{51} &  89.88\\ 
         &  & \ding{51} & \ding{55} &  88.55\\ 
         &  & \ding{51} & \ding{51} &  91.23\\ 
         \hline
        \multirow{4}{*}{4} & \multirow{4}{*}{1.4833} & \ding{55} & \ding{55} & 83.58 \\ 
         &  & \ding{55} & \ding{51} & 89.17 \\ 
         &  & \ding{51} & \ding{55} & 85.07 \\ 
         &  & \ding{51} & \ding{51} & 90.30 \\ 
         \hline
        \multirow{4}{*}{6} & \multirow{4}{*}{1.8169} & \ding{55} & \ding{55} & 78.71 \\ 
         &  & \ding{55} & \ding{51} &  89.29\\ 
         &  & \ding{51} & \ding{55} &  80.46\\ 
         &  & \ding{51} & \ding{51} &  89.65\\ 
         \hline
    \end{tabular}
    \caption{Impact of the Set Abstraction (SA) module and AMP2 components on the classification accuracy of Spiking PointNet++ (SSG) for ModelNet40. AWP: Activation-wise Propagation; RMP: Residual Membrane Potential connection.}

    \label{tab:abl_point}
\end{table}

In \tabref{tab:abl_point}, we conduct an ablation study on the two components of AMP2: Activation-wise Propagation (AWP) and Residual Membrane Potential Connection (RMP). Additionally, to assess the impact of network depth on AMP2, we vary the number of Set Abstraction (SA) modules in Spiking PointNet++ (SSG) to 3 (baseline), 4, and 6. We observe a decreasing trend in performance as the number of SA blocks increases, attributed to the absence of explicit residual connections in Spiking PointNet++. When equipped with either AWP or RMP alone, consistent accuracy improvements are observed on ModelNet40.

We further evaluate AMP2 on transformer-based SNN, including Spike-Driven Transformer V2 (SDT-v2) \cite{yao2024spikedriven} and Spikformer \cite{zhou2023spikformer} with default surrogates on an 100-category subset of ImageNet. Results in \tabref{tab:abl_transformer} confirm the effectiveness of AMP2 in enhancing performance. Particularly, removing AWP while keeping only RMP causes extremely slow loss convergence on SDT-v2, indicating the transformer architecture's reliance on AWP.

\begin{table}[!htb]
    \centering
    \begin{tabular}{lllccc}
    \hline
        \multirow{2}{*}{Model} &\multirow{2}{*}{Param(M)} & \multirow{2}{*}{T} & \multicolumn{2}{c}{AMP2} & \multirow{2}{*}{Acc.(\%)} \\ 
        &  & & AWP & RMP & \\ 
        \hline
        \multirow{4}{*}{SDT-v2} & \multirow{4}{*}{15} & \multirow{4}{*}{1} & \ding{55} & \ding{55} & 80.91 \\ 
         &  &  & \ding{55} & \ding{51} & N/A \\ 
         &  &  & \ding{51} & \ding{55} & 80.08 \\ 
         &  &  & \ding{51} & \ding{51} & 80.96 \\ 
         \hline
        \multirow{2}{*}{Spikformer} & \multirow{2}{*}{29.68} & 1 & \ding{55} & \ding{55} & 78.0 \\ 
         &  & 4 & \ding{55} & \ding{55} & 80.36 \\ \hline
    \end{tabular}
    \caption{Performance of transformer-based SNN on ImageNet100. SDT-v2 refers to Spike-Driven Transformer V2 (8-256), and Spikformer denotes the 8-512 variant. T indicates timesteps.}
    \label{tab:abl_transformer}
\end{table}
\subsection{Training Curves}
\begin{figure}[!htb]
    \centering
    \includegraphics[width=0.95\columnwidth]{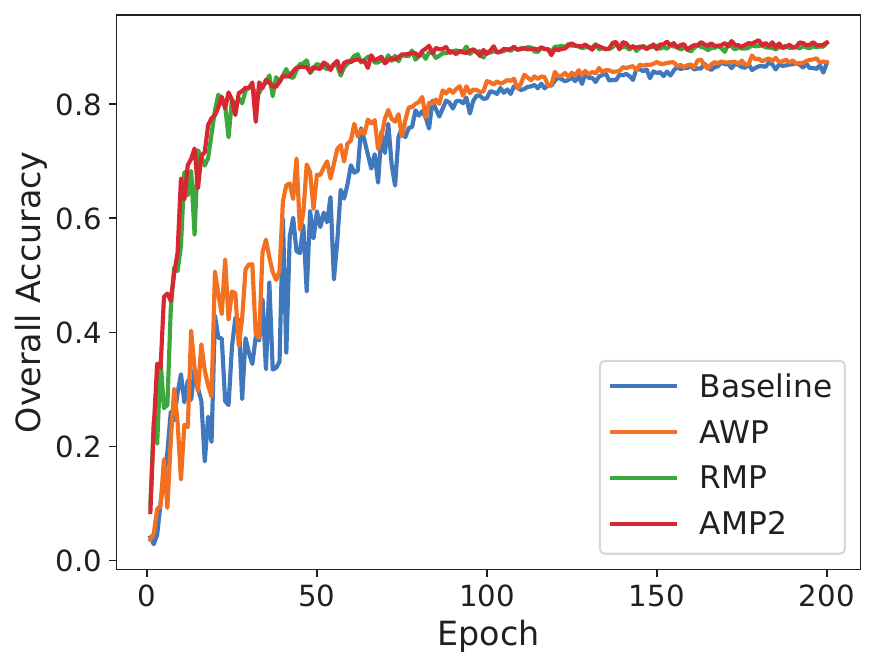}
    \caption{Training curves of Spiking PointNet++ (SSG) with varying configuration on ModelNet40.}
    \label{fig:curve}
\end{figure}

\figref{fig:curve} demonstrates the influence of AMP2 and its components AWP and RMP on the training of Spiking PointNet++ (SSG). Compared with the baseline trained with a single timestep, AWP yields consistent improvements across nearly all epochs. Similarly, the complete AMP2 configuration achieves faster accuracy gains than RMP during the first 50 epochs.
\subsection{Firing Rate}
Apart from accuracy, firing rate serves as a key indicator of energy efficiency in SNNs. As shown in \tabref{tab:fr}, we report the spike firing rate of all activation layers within Spiking PointNet++. Each SA module contains three spiking activation layers, we compute the average firing rate across them to represent the module's overall spiking activity. A consistent decline in firing rate across the three SA modules suggests progressively more efficient feature extraction. In contrast, the classification head exhibits a higher firing rate, indicating increased energy use for recognition—an intentional design that prioritizes task-critical computation while reducing overall overhead.
\begin{table}[!htb]
    \centering
    \begin{tabular}{ccc}
    \hline
        \multirow{2}{*}{Spiking PointNet++} & \multicolumn{2}{c}{Firing Rate ($\downarrow$)}   \\ 
         & w/o AMP2 & w/ AMP2 \\ 
         \hline
        Cls. LIF1 & \textbf{23.78} & 26.11 \\ 
        Cls. LIF2 & \textbf{33.78} & 36.10 \\ 
        SA1 LIF Avg. & 31.44 & \textbf{24.70} \\ 
        SA2 LIF Avg. & 27.23 & \textbf{22.35} \\ 
        SA3 LIF Avg. & 22.29 & \textbf{18.23} \\ 
        \hline
    \end{tabular}
    \caption{Firing rate of Spiking PointNet++ (SSG) in each spiking activation layer.}
    \label{tab:fr}
\end{table}

\section{Conclusion And Discussion}
In this paper, we propose \textbf{A}ctivation-wise \textbf{M}embrane \textbf{P}otential \textbf{P}ropagtion (AMP2) to enable single-timestep training and inference for SNNs. AMP2 is comprised of Activation/Neuron-wise Propagation (AWP) and Residual Membrane Potential Connection (RMP). AWP is a brain-inspired strategy that replaces timestep-wise membrane potential accumulation with accumulation across spatially adjacent neurons. RMP introduces a genuine fusion for membrane potential in SNN.

\textbf{Future work}. This study leaves two directions for future exploration. First, given sufficient computational resources, it is worth evaluating AMP2 on large-scale datasets such as ImageNet-1k and ES-ImageNet. Second, collaboration with hardware team is encouraged, as current neuromorphic hardware still lacks basic capabilities, like precise power measurement for SNN. We hope AMP2 can bridge the gap between efficient SNN and real-world applications.

\section{Acknowledgments}
This work was supported by the National Science and Technology Innovation 2030 - Major Project (Grant No. 2022ZD0208800), and NSFC General Program (Grant No. 62176215 and Grant No. 62573362)


\bibliography{aaai2026}

@InProceedings{Qi_2017_CVPR,
author = {Qi, Charles R. and Su, Hao and Mo, Kaichun and Guibas, Leonidas J.},
title = {PointNet: Deep Learning on Point Sets for 3D Classification and Segmentation},
booktitle = {Proceedings of the IEEE Conference on Computer Vision and Pattern Recognition (CVPR)},
month = {July},
year = {2017}
}

@article{qi2017pointnet++,
  title={Pointnet++: Deep hierarchical feature learning on point sets in a metric space},
  author={Qi, Charles Ruizhongtai and Yi, Li and Su, Hao and Guibas, Leonidas J},
  journal={Advances in neural information processing systems},
  volume={30},
  year={2017}
}

@inproceedings{wu20153d,
  title={3d shapenets: A deep representation for volumetric shapes},
  author={Wu, Zhirong and Song, Shuran and Khosla, Aditya and Yu, Fisher and Zhang, Linguang and Tang, Xiaoou and Xiao, Jianxiong},
  booktitle={Proceedings of the IEEE conference on computer vision and pattern recognition},
  pages={1912--1920},
  year={2015}
}

@inproceedings{yao2023spikedriven,
title={Spike-driven Transformer},
author={Man Yao and JiaKui Hu and Zhaokun Zhou and Li Yuan and Yonghong Tian and Bo XU and Guoqi Li},
booktitle={Thirty-seventh Conference on Neural Information Processing Systems},
year={2023},
url={https://openreview.net/forum?id=9FmolyOHi5}
}

@inproceedings{yao2024spikedriven,
title={Spike-driven Transformer V2: Meta Spiking Neural Network Architecture Inspiring the Design of Next-generation Neuromorphic Chips},
author={Man Yao and JiaKui Hu and Tianxiang Hu and Yifan Xu and Zhaokun Zhou and Yonghong Tian and Bo XU and Guoqi Li},
booktitle={The Twelfth International Conference on Learning Representations},
year={2024},
url={https://openreview.net/forum?id=1SIBN5Xyw7}
}

@article{yao2024scaling,
  title={Scaling Spike-driven Transformer with Efficient Spike Firing Approximation Training},
  author={Yao, Man and Qiu, Xuerui and Hu, Tianxiang and Hu, Jiakui and Chou, Yuhong and Tian, Keyu and Liao, Jianxing and Leng, Luziwei and Xu, Bo and Li, Guoqi},
  journal={arXiv preprint arXiv:2411.16061},
  year={2024}
}

@inproceedings{
zhou2023spikformer,
title={Spikformer: When Spiking Neural Network Meets Transformer },
author={Zhaokun Zhou and Yuesheng Zhu and Chao He and Yaowei Wang and Shuicheng YAN and Yonghong Tian and Li Yuan},
booktitle={The Eleventh International Conference on Learning Representations },
year={2023},
url={https://openreview.net/forum?id=frE4fUwz_h}
}

@article{zhou2024spikformer,
  title={Spikformer v2: Join the high accuracy club on imagenet with an snn ticket},
  author={Zhou, Zhaokun and Che, Kaiwei and Fang, Wei and Tian, Keyu and Zhu, Yuesheng and Yan, Shuicheng and Tian, Yonghong and Yuan, Li},
  journal={arXiv preprint arXiv:2401.02020},
  year={2024}
}

@ARTICLE{MS_ResNet,
  author={Hu, Yifan and Deng, Lei and Wu, Yujie and Yao, Man and Li, Guoqi},
  journal={IEEE Transactions on Neural Networks and Learning Systems}, 
  title={Advancing Spiking Neural Networks Toward Deep Residual Learning}, 
  year={2024},
  volume={},
  number={},
  pages={1-15},
  doi={10.1109/TNNLS.2024.3355393}}

@inproceedings{qiu2024efficient,
  title={Efficient 3D Recognition with Event-driven Spike Sparse Convolution},
  author={Qiu, Xuerui and Yao, Man and Zhang, Jieyuan and Chou, Yuhong and Qiao, Ning and Zhou, Shibo and Xu, Bo and Li, Guoqi},
  booktitle={Proceedings of the Thirty-Ninth AAAI Conference on Artificial Intelligence},
  year={2025}
}

@inproceedings{lan2023efficient,
  title={Efficient converted spiking neural network for 3d and 2d classification},
  author={Lan, Yuxiang and Zhang, Yachao and Ma, Xu and Qu, Yanyun and Fu, Yun},
  booktitle={Proceedings of the IEEE/CVF International Conference on Computer Vision},
  pages={9211--9220},
  year={2023}
}

@inproceedings{wu2024point,
  title={Point-to-Spike Residual Learning for Energy-Efficient 3D Point Cloud Classification},
  author={Wu, Qiaoyun and Zhang, Quanxiao and Tan, Chunyu and Zhou, Yun and Sun, Changyin},
  booktitle={Proceedings of the AAAI Conference on Artificial Intelligence},
  volume={38},
  number={6},
  pages={6092--6099},
  year={2024}
}

@article{ren2023spiking,
  title={Spiking pointnet: Spiking neural networks for point clouds},
  author={Ren, Dayong and Ma, Zhe and Chen, Yuanpei and Peng, Weihang and Liu, Xiaode and Zhang, Yuhan and Guo, Yufei},
  journal={Advances in Neural Information Processing Systems},
  volume={36},
  pages={41797--41808},
  year={2023}
}

@inproceedings{horowitz20141,
  title={1.1 computing's energy problem (and what we can do about it)},
  author={Horowitz, Mark},
  booktitle={2014 IEEE international solid-state circuits conference digest of technical papers (ISSCC)},
  pages={10--14},
  year={2014},
  organization={IEEE}
}

@article{ren2023spikepoint,
  title={Spikepoint: An efficient point-based spiking neural network for event cameras action recognition},
  author={Ren, Hongwei and Zhou, Yue and Huang, Yulong and Fu, Haotian and Lin, Xiaopeng and Song, Jie and Cheng, Bojun},
  journal={arXiv preprint arXiv:2310.07189},
  year={2023}
}

@inproceedings{wang2019space,
  title={Space-time event clouds for gesture recognition: From RGB cameras to event cameras},
  author={Wang, Qinyi and Zhang, Yexin and Yuan, Junsong and Lu, Yilong},
  booktitle={2019 IEEE Winter Conference on Applications of Computer Vision (WACV)},
  pages={1826--1835},
  year={2019},
  organization={IEEE},
  url={https://github.com/qwang014/EVclouds_gesture_recognition}
}

@inproceedings{thomas2019kpconv,
  title={Kpconv: Flexible and deformable convolution for point clouds},
  author={Thomas, Hugues and Qi, Charles R and Deschaud, Jean-Emmanuel and Marcotegui, Beatriz and Goulette, Fran{\c{c}}ois and Guibas, Leonidas J},
  booktitle={Proceedings of the IEEE/CVF international conference on computer vision},
  pages={6411--6420},
  year={2019}
}

@article{fang2021deep,
  title={Deep residual learning in spiking neural networks},
  author={Fang, Wei and Yu, Zhaofei and Chen, Yanqi and Huang, Tiejun and Masquelier, Timoth{\'e}e and Tian, Yonghong},
  journal={Advances in Neural Information Processing Systems},
  volume={34},
  pages={21056--21069},
  year={2021}
}

@misc{imagenet100_kaggle,
  title        = {ImageNet100},
  author       = {Ambityga},
  year         = {2023},
  howpublished = {\url{https://www.kaggle.com/datasets/ambityga/imagenet100}},
}

@article{ding2025rethinking,
  title={Rethinking spiking neural networks from an ensemble learning perspective},
  author={Ding, Yongqi and Zuo, Lin and Jing, Mengmeng and He, Pei and Deng, Hanpu},
  journal={arXiv preprint arXiv:2502.14218},
  year={2025}
}

@inproceedings{wu2025spiking,
  title={Spiking point transformer for point cloud classification},
  author={Wu, Peixi and Chai, Bosong and Li, Hebei and Zheng, Menghua and Peng, Yansong and Wang, Zeyu and Nie, Xuan and Zhang, Yueyi and Sun, Xiaoyan},
  booktitle={Proceedings of the AAAI Conference on Artificial Intelligence},
  volume={39},
  number={20},
  pages={21563--21571},
  year={2025}
}

@inproceedings{ding2024shrinking,
  title={Shrinking your timestep: Towards low-latency neuromorphic object recognition with spiking neural networks},
  author={Ding, Yongqi and Zuo, Lin and Jing, Mengmeng and He, Pei and Xiao, Yongjun},
  booktitle={Proceedings of the AAAI Conference on Artificial Intelligence},
  volume={38},
  number={10},
  pages={11811--11819},
  year={2024}
}

@article{zuo2024temporal,
  title={Temporal reversed training for spiking neural networks with generalized spatio-temporal representation},
  author={Zuo, Lin and Ding, Yongqi and Luo, Wenwei and Jing, Mengmeng and Tian, Xianlong and Yang, Kunshan},
  journal={arXiv e-prints},
  pages={arXiv--2408},
  year={2024}
}

@inproceedings{anumasa2024enhancing,
  title={Enhancing training of spiking neural network with stochastic latency},
  author={Anumasa, Srinivas and Mukhoty, Bhaskar and Bojkovic, Velibor and De Masi, Giulia and Xiong, Huan and Gu, Bin},
  booktitle={Proceedings of the AAAI Conference on Artificial Intelligence},
  volume={38},
  number={10},
  pages={10900--10908},
  year={2024}
}

@inproceedings{shi2024spikingresformer,
  title={Spikingresformer: Bridging resnet and vision transformer in spiking neural networks},
  author={Shi, Xinyu and Hao, Zecheng and Yu, Zhaofei},
  booktitle={Proceedings of the IEEE/CVF conference on computer vision and pattern recognition},
  pages={5610--5619},
  year={2024}
}

@article{zhou2024qkformer,
  title={Qkformer: Hierarchical spiking transformer using qk attention},
  author={Zhou, Chenlin and Zhang, Han and Zhou, Zhaokun and Yu, Liutao and Huang, Liwei and Fan, Xiaopeng and Yuan, Li and Ma, Zhengyu and Zhou, Huihui and Tian, Yonghong},
  journal={Advances in Neural Information Processing Systems},
  volume={37},
  pages={13074--13098},
  year={2024}
}

@inproceedings{zhao2021point,
  title={Point transformer},
  author={Zhao, Hengshuang and Jiang, Li and Jia, Jiaya and Torr, Philip HS and Koltun, Vladlen},
  booktitle={Proceedings of the IEEE/CVF international conference on computer vision},
  pages={16259--16268},
  year={2021}
}

@inproceedings{uy2019revisiting,
  title={Revisiting point cloud classification: A new benchmark dataset and classification model on real-world data},
  author={Uy, Mikaela Angelina and Pham, Quang-Hieu and Hua, Binh-Son and Nguyen, Thanh and Yeung, Sai-Kit},
  booktitle={Proceedings of the IEEE/CVF international conference on computer vision},
  pages={1588--1597},
  year={2019}
}

@inproceedings{zheng2021going,
  title={Going deeper with directly-trained larger spiking neural networks},
  author={Zheng, Hanle and Wu, Yujie and Deng, Lei and Hu, Yifan and Li, Guoqi},
  booktitle={Proceedings of the AAAI conference on artificial intelligence},
  volume={35},
  number={12},
  pages={11062--11070},
  year={2021}
}

@INPROCEEDINGS{8100264,
  author={Amir, Arnon and Taba, Brian and Berg, David and Melano, Timothy and McKinstry, Jeffrey and Di Nolfo, Carmelo and Nayak, Tapan and Andreopoulos, Alexander and Garreau, Guillaume and Mendoza, Marcela and Kusnitz, Jeff and Debole, Michael and Esser, Steve and Delbruck, Tobi and Flickner, Myron and Modha, Dharmendra},
  booktitle={2017 IEEE Conference on Computer Vision and Pattern Recognition (CVPR)}, 
  title={A Low Power, Fully Event-Based Gesture Recognition System}, 
  year={2017},
  volume={},
  number={},
  pages={7388-7397},
  doi={10.1109/CVPR.2017.781}}

@article{10.1145/2980179.2980238,
author = {Yi, Li and Kim, Vladimir G. and Ceylan, Duygu and Shen, I-Chao and Yan, Mengyan and Su, Hao and Lu, Cewu and Huang, Qixing and Sheffer, Alla and Guibas, Leonidas},
title = {A scalable active framework for region annotation in 3D shape collections},
year = {2016},
issue_date = {November 2016},
publisher = {Association for Computing Machinery},
address = {New York, NY, USA},
volume = {35},
number = {6},
issn = {0730-0301},
url = {https://doi.org/10.1145/2980179.2980238},
doi = {10.1145/2980179.2980238},
journal = {ACM Trans. Graph.},
month = dec,
articleno = {210},
numpages = {12}
}

@inproceedings{armeni20163d,
  title={3d semantic parsing of large-scale indoor spaces},
  author={Armeni, Iro and Sener, Ozan and Zamir, Amir R and Jiang, Helen and Brilakis, Ioannis and Fischer, Martin and Savarese, Silvio},
  booktitle={Proceedings of the IEEE conference on computer vision and pattern recognition},
  pages={1534--1543},
  year={2016}
}

@article{fang2023spikingjelly,
  title={Spikingjelly: An open-source machine learning infrastructure platform for spike-based intelligence},
  author={Fang, Wei and Chen, Yanqi and Ding, Jianhao and Yu, Zhaofei and Masquelier, Timoth{\'e}e and Chen, Ding and Huang, Liwei and Zhou, Huihui and Li, Guoqi and Tian, Yonghong},
  journal={Science Advances},
  volume={9},
  number={40},
  pages={eadi1480},
  year={2023},
  publisher={American Association for the Advancement of Science}
}

\appendix
\section{Datasets}
\textbf{ModelNet40} is a widely-used benchmark dataset for 3D shape classification. It is a 40-Class subset containing synthetic object point clouds from CAD models. All models are aligned and cleaned for training and testing. Performance is evaluated by overall accuracy (OA) and mean class accuracy (mAcc).

\textbf{ScanObjectNN} is a more challenging real-world 3D object classification benchmark derived from scanned indoor scenes. Unlike the well-structured ModelNet, ScanObjectNN includes background clutter, occlusion, and misalignments.

\textbf{DVS128Gesture} dataset is a neuromorphic gesture recognition benchmark captured using a Dynamic Vision Sensor (DVS). It comprises 11 gesture categories, with each event sample represented as $[t, p, x, y]$ at a $128 \times 128$ resolution. The dataset includes approximately 1,400 recordings from 29 subjects across three illumination conditions, establishing it as a key benchmark for SNN research.

\textbf{ShapeNetPart } is a widely used benchmark dataset for 3D point cloud part segmentation. It is a subset of the ShapeNet dataset and contains 16 object categories, with each sample annotated with per-point part label across 50 distinct part, enabling fine-grained semantic segmentation in point-based deep learning.

\textbf{S3DIS} contains 6 large-scale indoor areas collected from Stanford campus buildings, comprising over 270 rooms across various functional spaces such as offices, conference rooms, and hallways. Each point in the dataset is annotated with semantic labels from 13 object categories

\section{Experiment}
\subsection{Pre-Processing}
For point cloud data preprocessing, we mainly follow the approach outlined in \citet{ren2023spiking}. For neuromorphic data, we use the sliding window technique from \citet{wang2019space} to obtain point-based DVS data. We leverage SpikingJelly \cite{fang2023spikingjelly} to integrate events into $T$ event frames, where $T$ corresponds to the timestep of the SNN. Finally, the event frames are downsampled to a resolution of $48 \times 48$ before being input to the SNN.
\subsection{Hyperparameter}
We use the same hyperparameters as \citet{ren2023spiking}, as shown in \tabref{tab:hyper}. All other training details follow the original papers, including Spiking PointNet \cite{ren2023spiking} and SEW ResNet \cite{fang2021deep}.
\begin{table}[]
    \centering
    \begin{tabular}{ccccc}
    \hline
         $k$ & $c$ & $\beta$ & $\alpha$ &$V_{th}$ \\
         \hline
         5 & 0.5 & 0.25 & 0.8 & 1.0 \\
         \hline
    \end{tabular}
    \caption{Hyperparameters used in AMP2 and Spiking PointNet}
    \label{tab:hyper}
\end{table}

\subsection{Hardware}
All experiments were conducted with the PyTorch package. Most models were trained on NVIDIA V100 except our self-reproduced transformer SNN on A100. The random seed is set to 42. Optimizer used is Adam with learning rate starting at $1e4$. Most settings are the same as those of backbone models.
\section{Pseudocode for MP Initialization of AMP2}
\begin{algorithm}[htb]
   \caption{Spiking mechanism of the $i$th AMP2-based LIF neuron layer}
\begin{algorithmic}
   \STATE {\bfseries Input:} Input $X_i$, residual $MP_{i-1}$, spiking threshold $V_{th}$, decay coefficient $\beta$, history coefficients $\alpha$, random initialized $MP_{random}\in({0,1})$
   \IF{$MP_{i-1}$ is None}
      \STATE $MP_{i}^{0} \leftarrow$ $MP_{random}$
   \ELSE
      \STATE $MP_{i}^{0} \leftarrow \alpha \cdot MP_{i-1}^1 + (1-\alpha) \cdot$ $MP_{random}$
   \ENDIF
   \STATE $MP_{i}^1 \leftarrow \beta \cdot MP_{i}^{0} + X_i$
   \STATE $S_{i} \leftarrow \text{Hea.}(\frac{MP_l^1}{V_{th}})$
   \STATE $MP_i^{2} \leftarrow MP_i^1(1-S_i)$ 
   \COMMENT{$MP^1$ is passed to next neuron layer}
   \STATE {\bfseries Output:} $S_{i}$
\end{algorithmic}
\label{algo:amp2}
\end{algorithm}

\section{Parameters Count}
We summarize the number of parameters for the majority of models used in this paper in \tabref{tab:parameter}.
\begin{table}[htbp]
    \centering
    \begin{tabular}{ll}
    \hline
        Model & Parameters(M) \\ \hline
        P2SResLNet & $>$20$^{*}$ \\ 
        SpikingResformer-Ti & 11.14 \\ 
        Spiking PointNet & 3.46 \\ 
        Spiking PointNet++(SSG) & 1.48 \\ 
        Spiking PointNet++(MSG) & 1.74 \\ 
        ResNet-tiny & 0.13 \\ 
        Spiking Driven Transformer(SDT-v1,2-256) & 4.1$^{*}$\\
        Spiking Point Transformer(SPT) & 9.6 \\ 
        SpikePoint(DVS128) & 0.58 \\ 
        Efficient 3D SNN (E3DSNN)& 1.87\\
        \hline
    \end{tabular}
    \caption{Parameters comparison of models involved in experiments. $*$ denotes estimated values based on related models due to incomplete coverage in original papers.}
    \label{tab:parameter}
\end{table}

\section{Residual Connection}
\begin{figure}[htbp]
    \centering
    \includegraphics[width=\columnwidth]{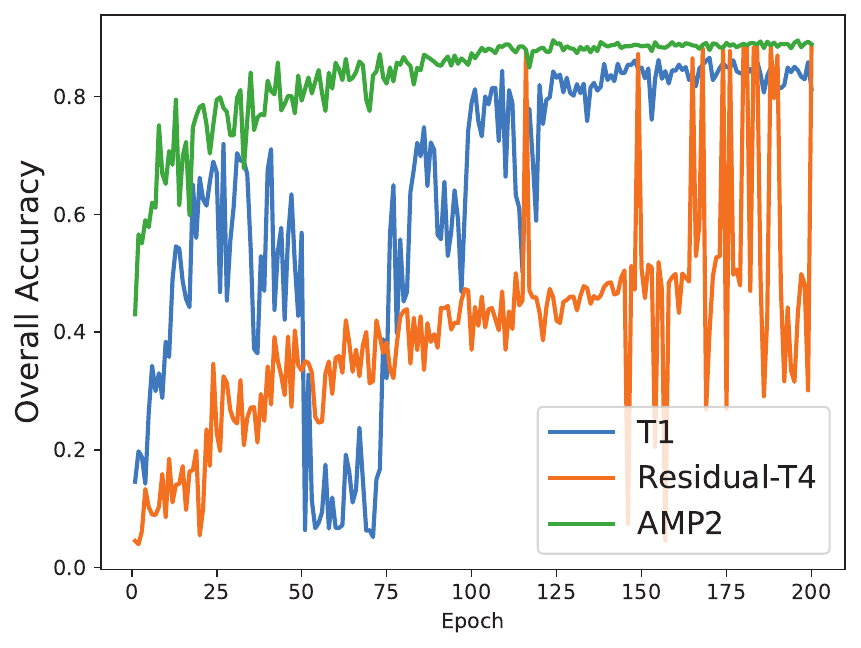}
    \caption{Training comparison of Membrane Shortcut (MS) and AMP2  in Spiking PointNet on ModelNet40. T1 represents a single-timestep SNN, and Residual-T4 refers to an SNN with MS trained under 4 timesteps.}
    \label{fig:resnet}
\end{figure}

\begin{figure}[htbp]
    \centering
    \includegraphics[width=\columnwidth]{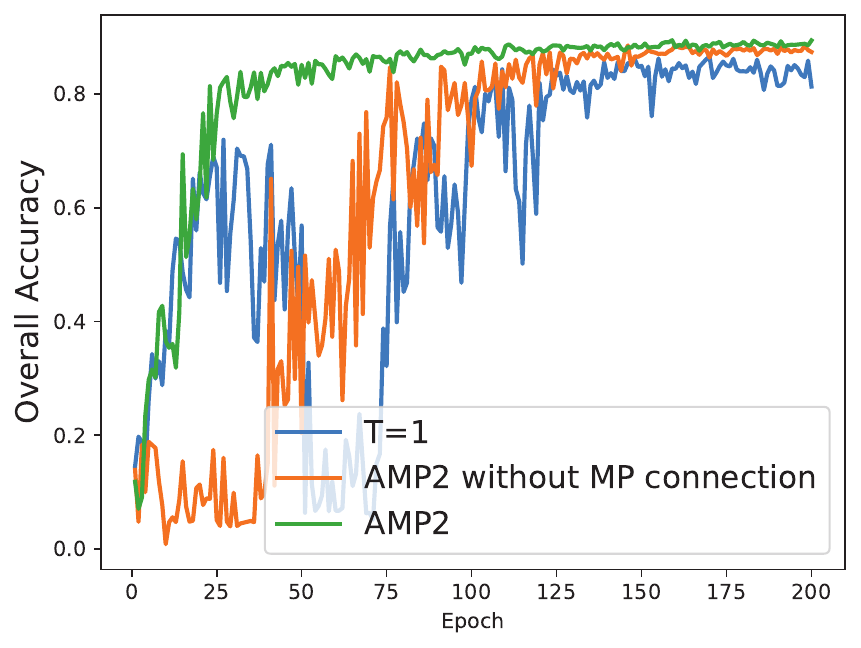}
    \caption{Training comparison of AWP (w/o MP Connection) and AMP2 in Spiking PointNet on ModelNet40. $T=1$ represents a single-timestep baseline.}
    \label{fig:no_RMP}
\end{figure}

We conduct two additional ablation studies on the RMP module. The first study compares RMP with the Membrane Shortcut (MS) in MS ResNet \cite{MS_ResNet} using the Spiking PointNet model, as shown in \figref{fig:resnet}. The second study ablates the RMP module in Spiking PointNet, as illustrated in \figref{fig:no_RMP}.

\end{document}